\documentclass{article}

\usepackage{times}
\usepackage{graphicx} 
\usepackage{wrapfig}
\usepackage{subfigure} 
\usepackage{multirow}
\usepackage{adjustbox}

\usepackage{natbib}

\usepackage{listings}

\usepackage{algorithm}
\usepackage{algorithmic}
\renewcommand{\algorithmicrequire}{\textbf{Input:}}
\renewcommand{\algorithmicensure}{\textbf{Output:}}

\newcommand{\X}{\mathbf{X}}
\newcommand{\E}{\mathbb{E}}
\newcommand{\x}{\mathbf{x}}
\newcommand{\z}{\mathbf{z}}

\usepackage{tikz}
\newcommand*\circled[1]{\tikz[baseline=(char.base)]{
            \node[shape=circle,draw,inner sep=2pt] (char) {#1};}}

\usepackage[accepted]{arxiv} 

\usepackage{amssymb}
\usepackage{amsmath}

\usepackage[T1]{fontenc}
\usepackage[latin1]{inputenc}
\usepackage{courier}

\usepackage{xspace}
\newcommand{\function}[1]{\ensuremath{\mathtt{#1}}\xspace}

\icmltitlerunning{Grammar Variational Autoencoder}

\begin{document} 

\twocolumn[
\icmltitle{Grammar Variational Autoencoder}



\icmlsetsymbol{equal}{*}

\begin{icmlauthorlist}
\icmlauthor{Matt J. Kusner}{ati,ww}
\icmlauthor{Brooks Paige}{ati,cam}
\icmlauthor{Jos\'e Miguel Hern\'andez-Lobato}{cam}

\end{icmlauthorlist}

\icmlaffiliation{ati}{Alan Turing Institute}
\icmlaffiliation{ww}{University of Warwick}
\icmlaffiliation{cam}{University of Cambridge}

\icmlcorrespondingauthor{}{mkusner@turing.ac.uk}
\icmlcorrespondingauthor{}{bpaige@turing.ac.uk}
\icmlcorrespondingauthor{}{jmh233@cam.ac.uk}

\icmlkeywords{context-free grammar, variational autoencoder}

\vskip 0.3in
]



\printAffiliationsAndNotice{}  

\begin{abstract}
Deep generative models have been wildly successful at learning coherent latent representations for continuous data such as video and audio. However, generative modeling of discrete data such as arithmetic expressions and molecular structures still poses significant challenges. Crucially, state-of-the-art methods often produce outputs that are not valid. We make the key observation that frequently, discrete data can be represented as a parse tree from a context-free grammar. We propose a variational autoencoder which encodes and decodes directly to and from these parse trees, ensuring the generated outputs are always valid. Surprisingly, we show that not only does our model more often generate valid outputs, it also learns a more coherent latent space in which nearby points decode to similar discrete outputs. We demonstrate the effectiveness of our learned models by showing their improved performance in Bayesian optimization for symbolic regression and molecular synthesis.
\end{abstract} 

\section{Introduction}
Generative machine learning models have been used recently to produce extraordinary results, from realistic musical improvisation \citep{jaques2016tuning}, to changing facial expressions in images \citep{radford2015unsupervised,upchurch2016deep}, to creating realistic looking artwork \citep{gatys2015neural}. 
%
In large part, these generative models have been successful at representing data in continuous domains. Recently there is increased interest in training generative models to construct more complex, discrete data types such as arithmetic expressions \citep{kusner2016gans}, 
source code \citep{gaunt2016terpret,riedel2016programming} and molecules \citep{gomez2016automatic}.

To train generative models for these tasks, these objects are often first represented as strings. This is in large part due to the fact that there exist powerful models for text sequence modeling such as Long Short Term Memory networks (LSTMs) \citep{hochreiter1997long}, Gated Recurrent Units (GRUs) \citep{cho2014learning}, and Dynamic Convolutional Neural Networks (DCNNs) \citep{kalchbrenner2014convolutional}. For instance, molecules can be represented by so-called SMILES strings \citep{weininger1988smiles} and \citet{gomez2016automatic} has recently developed a generative model for molecules based on SMILES strings that uses GRUs and DCNNs. This model is able to encode and decode molecules to and from a continuous latent space, allowing one to search this space for new molecules with desirable properties \citep{gomez2016automatic}.

However, one immediate difficulty in using strings to represent molecules is that the representation is very brittle: small changes in the string can lead to completely different molecules, or often do not correspond to valid molecules at all. Specifically, \citet{gomez2016automatic} described that while searching for new molecules, the probabilistic decoder --- the distribution which maps 
from the continuous latent space into the space of molecular structures --- would sometimes accidentally put high probability on strings which are not valid SMILES strings or do not encode plausible molecules.

To address this issue, we propose to directly incorporate knowledge about the structure of discrete data using a \emph{grammar}. Grammars exist for a wide variety of discrete domains such as symbolic expressions \citep{allamanis2016learning},
standard programming languages such as C \citep{kernighan1988c}, 
and chemical structures \citep{james2015opensmiles}. For instance the set of syntactically valid \textsc{SMILES} strings is described using a
context free grammar, which can be used for parsing and validation\footnote{http://opensmiles.org/spec/open-smiles-2-grammar.html}. 

Given a grammar, every valid discrete object can be described as a parse tree from the grammar. Thus, we propose the \emph{grammar variational autoencoder} (GVAE) which encodes and decodes directly to and from these parse trees. Generating parse trees as opposed to text ensures that all outputs are valid based on the grammar. This frees the GVAE from learning syntactic rules and allows it to wholly focus on learning other `semantic' properties.


We demonstrate the GVAE on two different tasks for generating discrete data: 1)
generating simple arithmetic expressions and 2) generating valid molecules. We
show not only does our model produce a higher proportion of valid discrete outputs than a
character based autoencoder, it also produces smoother latent representations.
We also show that this learned latent space is effective for searching for
arithmetic expressions that fit data, for finding better drug-like molecules,
and for making accurate predictions about target properties.



\section{Background}
\label{sec:background}

\subsection{Variational autoencoder}

We wish to learn both an encoder and a decoder
for mapping data $\x$ to and from values $\z$ in a continuous space.
The variational autoencoder \cite{kingma2014auto,rezende2014stochastic} provides a
formulation in which the encoding $\z$ is interpreted as a latent variable in a probabilistic
generative model; a probabilistic decoder is defined by a likelihood function $p_\theta(\x | \z)$
and parameterized by $\theta$.
Alongside a prior distribution $p(\z)$ over the latent variables, the posterior distribution
$p_\theta(\z | \x) \propto p(\z)p_\theta(\x|\z)$ can then be interpreted as a probabilistic encoder.

To admit efficient inference, the variational Bayes approach simultaneously learns
both the parameters of $p_\theta(\x|\z)$ as well as those of a posterior approximation
$q_\phi(\z | \x)$.
This is achieved by maximizing the evidence lower bound (ELBO)
\begin{align}
\mathcal{L}(\phi, \theta; \x)
&= 
\E_{q(\z|\x)}\left[ \log p_\theta(\x,\z) - \log q_\phi(\z|\x) \right],
\label{eq:elbo}
\end{align}
with $\mathcal{L}(\phi, \theta; \x) \leq \log p_\theta(\x)$. So long as $p_\theta(\x|\z)$ and $q_\phi(\z|\x)$ can be computed pointwise, and are
differentiable with respect to their parameters, the ELBO can be maximized via gradient descent;
this allows wide flexibility in choice of encoder and decoder models.
Typically these will take the form of exponential family distributions whose parameters are the
output of a multi-layer neural network.


\subsection{Context-free grammars}

A context-free grammar (CFG) is traditionally defined as a 4-tuple $G = (V, \Sigma, R, S)$:
$V$ is a finite set of non-terminal symbols;
the {\em alphabet} $\Sigma$ is a finite set of terminal symbols, disjoint from $V$;
$R$ is a finite set of production rules;
and $S$ is a distinct non-terminal known as the {\em start symbol}.
The rules $R$ are formally described as $\alpha \rightarrow \beta$ for $\alpha \in V$ and $\beta \in (V \cup \Sigma)^*$,
with $^*$ denoting the Kleene closure. In practice, these rules are defined as a set of mappings 
from a single left-hand side non-terminal in $V$ to a sequence of terminal and/or non-terminal symbols,
and can be interpreted as a rewrite rule.

Application of a production rule to a non-terminal symbol defines a tree, with
symbols on the right-hand side of the production rule becoming child nodes for the left-hand side parent.
The grammar $G$ thus defines a set of possible trees extending from each non-terminal symbol in $V$,
produced by recursively applying rules in $R$ to leaf nodes until all leaf nodes are terminal symbols in $\Sigma$.
The {\em language} of $G$ is set of all sequences of terminal symbols which can be produced by a 
left-to-right traversal of the leaf nodes in a tree.
Given a string in the language (i.e., a sequence of terminals), 
a {\em parse tree} is a tree rooted at $S$ which has this sequence of terminal symbols as its leaf nodes.
The ubiquity of context-free languages in computer science is due in part to the presence of efficient parsing algorithms.
For more background on context free grammars and automata theory, see e.g.~\citet{hopcroft2006automata}.

The context-free grammar can form the backbone of a probabilistic generative model for valid strings.
By assigning probabilities to each production rule in the grammar, it is possible to define a probability distribution
over parse trees \cite{baker1979trainable,booth1973applying}.
A string can be generated by repeatedly sampling and applying production rules, beginning from the start symbol, until no non-terminals remain.
Modern approaches allow the probabilities used to at each stage to depend on the current state of the parse tree \citep{johnson2007adaptor}.


\begin{figure*}[t]
\begin{center}
\centerline{\includegraphics[width=\textwidth]{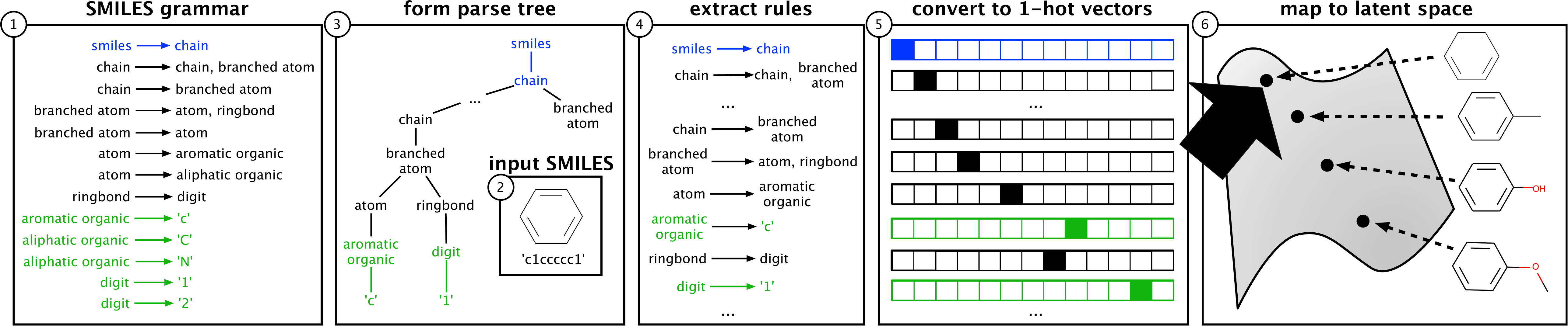}}
\vspace{-2ex}
\caption{The encoder of the GVAE. We denote the start rule in blue and all rules that decode to terminal in green. See text for details. \label{figure.encoding}}
\vspace{-2ex}
\end{center}
\end{figure*}

\section{Methods}
\label{sec:method}

In this section we describe how a grammar can improve variational autoencoders (VAE) for discrete data. 
It will do so by drastically reducing the number of invalid outputs generated from the VAE. 

One glaring issue with the character VAE is that it may frequently map latent points to sequences that are not valid, 
hoping the VAE will infer from training data what constitutes a valid sequence.
Instead of implicitly encouraging the VAE to produce valid molecules, we propose to give the VAE explicit knowledge about how to produce valid molecules. We do this by using a grammar for the sequences: given a grammar we can take any valid sequence and parse it into a sequence of production rules. 
Applying these rules in order will yield the original sequence. Our approach will be to learn a VAE that produces sequences of grammar production rules. 
The benefit is that it is trivial to generate valid sequences of production rules, as the grammar describes the valid set of rules that can be selected at any point during the generation process. 
Thus our model is able to focus on learning semantic properties of sequence data without also having to learn syntactic constraints. We describe our model in detail on a small example.

\subsection{An illustrative example}
We propose a grammar variational autoencoder (GVAE) that encodes and decodes in the space of grammar production rules. We describe how the GVAE works using a simple example.

\paragraph{Encoding.}
Consider a subset of the SMILES grammar as shown in Figure~\ref{figure.encoding}, box \circled{1}. These are the possible production rules that can be used for constructing a molecule.
Imagine we are given as input the SMILES string for benzene: `c1ccccc1'. Figure~\ref{figure.encoding}, box \circled{2} shows this molecule. To encode this molecule into a continuous latent representation we begin by using the SMILES grammar to parse this string into a parse tree (partially shown in box \circled{3}). 
This tree describes how `c1ccccc1' is generated by the grammar. 
We decompose this tree into a sequence of production rules by performing a pre-order traversal on the branches of the parse tree going from left-to-right, shown in box \circled{4}. 
We convert these rules into 1-hot indicator vectors, where each dimension corresponds to a rule in the SMILES grammar, box \circled{5}. 
Letting $K$ denote the total number of production rules in the entire grammar, 
and $T(\X)$ the number of productions applied in total to generate the output string for $\X$,
the collection of 1-hot vectors can be written as a $T(\X) \times K$ matrix $\X$.
We use a deep convolutional neural network to map this collection of 1-hot vectors $\X$ to a continuous latent vector $\+z$ 
The architecture of the encoding network is described in the supplementary material.

\begin{figure*}[t]
\begin{center}
\vspace{-1ex}
\centerline{\includegraphics[width=\textwidth]{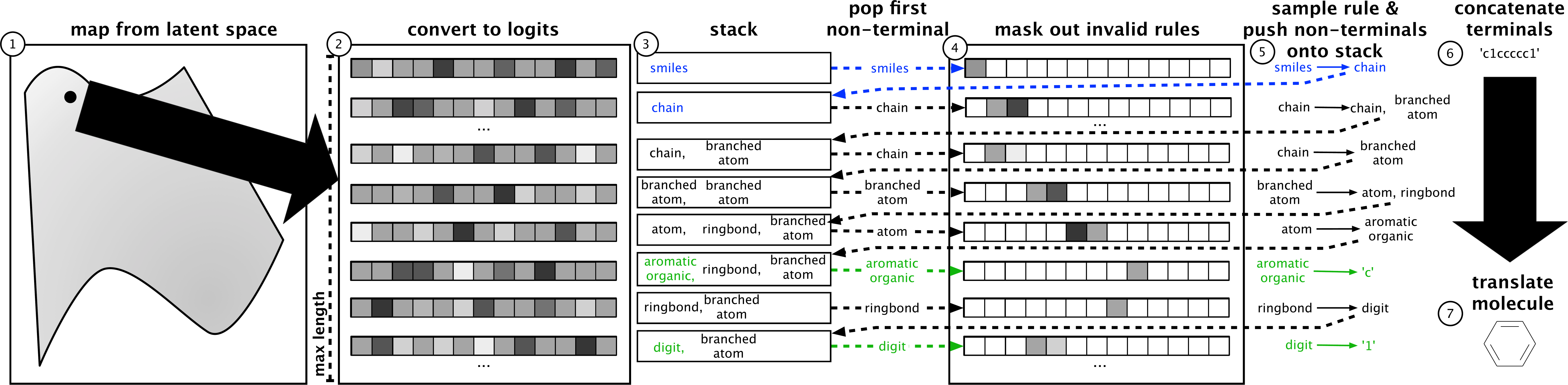}}
\vspace{-2ex}
\caption{The decoder of the GVAE. See text for details. \label{figure.decoding}}
\vspace{-2ex}
\end{center}
\end{figure*}

\paragraph{Decoding.}
We now describe how we map continuous vectors back to a sequence of production rules (and thus SMILES strings). Crucially we construct the decoder so that at any time while we are decoding this sequence the decoder will only be allowed to select a subset of production rules that are `valid'. This will cause the decoder to only produce valid parse sequences from the grammar.

We begin by passing the continuous vector $\+z$ through a recurrent neural network which produces a set of unnormalized log probability vectors (or `logits'), shown in Figure~\ref{figure.decoding}, box \circled{1} and \circled{2}. Exactly like the 1-hot vectors produced by the encoder, each dimension of the logit vectors corresponds to a production rule in the grammar. We can again write these collection of logit vectors as a matrix $\+F \in \mathbb{R}^{T_{max} \times K}$, where $T_{max}$ is the maximum number of timesteps (production rules) allowed by the decoder. We will use these vectors in the rest of the decoder to select production rules. 

To ensure that any sequence of production rules generated from the decoder is valid, we keep track of the state of the parsing using a last-in first-out (LIFO) stack. This is shown in Figure~\ref{figure.decoding}, box \circled{3}. At the beginning, every valid parse from the grammar must start with the start symbol: \emph{smiles}, which is placed on the stack. Next we pop off whatever non-terminal symbol that was placed last on the stack (in this case \emph{smiles}), and we use it to mask out the invalid dimensions of the logit vector. Formally, for every non-terminal $\alpha$ we define a fixed binary mask vector $\+m_\alpha \in [0,1]^K$. This takes the value `1' for all indices in $1,\ldots,K$ corresponding to production rules that have $\alpha$ on their left-hand-side.

In this case the only production rule in the grammar beginning with \emph{smiles} is the first so we zero-out every dimension except the first, shown in Figure~\ref{figure.decoding}, box \circled{4}. We then sample from the remaining unmasked rules, using their values in the logit vector. To sample from this masked logit at any timestep $t$ we form the following masked distribution:
\begin{align}
p(\x_{t} = k | \alpha, \+z) = \frac{ m_{\alpha,k} \exp(f_{tk})}{\sum_{j=1}^K m_{\alpha,k} \exp(f_{tj})},
\label{eq:prob-productions}
\end{align}
where $f_{tk}$ is the $(t,k)$-element of the logit matrix $\+F$. As only the first rule is unmasked we will select this rule \emph{smiles}~$\rightarrow$~\emph{chain} as the first rule in our generated sequence.

Now the next rule must begin with \emph{chain}, so we push it onto the stack (Figure~\ref{figure.decoding}, box \circled{3}). We sample this non-terminal and again use it to mask out all of the rules that cannot be applied in the current logit vector. We then sample a valid rule from this logit vector: \emph{chain}~$\rightarrow$~\emph{chain}, \emph{branched atom}. Just as before we push the non-terminals on the right-hand side of this rule onto the stack, adding the individual non-terminals in from right to left, such that the leftmost non-terminal is on the top of the stack. For the next state we again pop the last rule placed on the stack and mask the current logit, etc. This process continues until the stack is empty or we reach the maximum number of logit vectors $T_{max}$. We describe this decoding procedure formally in Algorithm 1. In practice, because sampling from the decoder often finishes before $t$ reaches $T_{max}$, we introduce an additional `no-op' rule to the grammar that we use to pad $\+X$ until the number of rows equals $T_{max}$.

We note the explicit connection between the process in Algorithm 1 and parsing algorithms for pushdown automata. 
A pushdown automaton is a finite
state machine which has access to a single stack for long-term storage, and are equivalent to context-free grammars
in the sense that every CFG can be converted into a pushdown automaton, and vice-versa \cite{hopcroft2006automata}.
The decoding algorithm performs the sequence of actions taken by a nondeterministic pushdown automaton at each
stage of a parsing algorithm; the nondeterminism is resolved by sampling according to the probabilities in the emitted logit vector.


\begin{algorithm}[t]
\begin{algorithmic}[1]
\REQUIRE Deterministic decoder output $\+F \in \mathbb{R}^{T_{max} \times K}$,
masks $\+m_\alpha$ for each production rule $\alpha$
\ENSURE Sampled productions $\+X$ from  $p(\+X | \+z)$
\STATE Initialize empty stack $\mathcal{S}$, and push the start symbol $S$ onto the top;
set $t=0$
\WHILE{$\mathcal{S}$ is nonempty}
\STATE Pop the last-pushed non-terminal $\alpha$ from the stack $\mathcal{S}$ 
\STATE Use Eq.~\eqref{eq:prob-productions} to sample a production rule $\mathcal{R}$ 
\STATE Set $\x_{t} \leftarrow \mathcal{R}$
\STATE Let $\textsc{RHS}(\mathcal{R})$ denote all non-terminals on the right-hand side of rule $\mathcal{R}$,
ordered from right to left
\FOR{ non-terminal $\beta$ in $\textsc{RHS}(\mathcal{R})$}
\STATE Push $\beta$ on to the stack $\mathcal{S}$
\ENDFOR
\STATE Set $\X \leftarrow [\X, \x_t]$
\STATE Set $t \leftarrow t+1$
\ENDWHILE
\caption{Sampling from the decoder}
\end{algorithmic}
\label{algo:sample-from-decoder}
\end{algorithm}

\paragraph{Contrasting the character VAE.}
Notice that the key difference between this grammar VAE decoder and a character-based VAE decoder is that at every point in the generated sequence, the character VAE can sample any possible character. There is no stack or masking operation. The grammar VAE however is constrained to select syntactically-valid sequences.

\paragraph{Syntactic vs. semantic validity.}
It is important to note that the grammar encodes \emph{syntactically valid} molecules but not necessarily \emph{semantically valid} molecules. This is because: 
1. certain molecules produced by the grammar may be very unstable molecules or not chemically-valid (for instance an oxygen atom cannot bond to 3 other atoms as it only has 2 free electrons for bonding, although it would be possible to generate this in a molecule from the grammar). 
2. The SMILES language has non-context free aspects such as a ringbond must be opened and closed by the same digit, starting with `1' (such is the case for benzene `c1ccccc1'). 
The particular challenge for matching digits, in contrast to matching grouping symbols such as parentheses, is that they do not compose in a nested manner; for example, `C12(CCCCC1)CCCCC2' is a valid SMILES string and molecule.
Furthermore, any intermediate ringbond must use digits that increment by one for each new ringbond. 
Keeping track of which digit to use for each ringbond is not context-free. 3. Finally, we note that the GVAE can output an undetermined sequence if there are still non-terminal symbols on the stack after processing all $T_{max}$ logit vectors. While this could be fixed by a procedure that converts these non-terminals to terminals, for simplicity we mark these sequences as invalid.


\subsection{Training}
During training, each input SMILES encoded as a sequence of 1-hot vectors $\+X \in \{0,1\}^{T_{max} \times K}$, also defines a sequence of $T_{max}$ mask vectors. 
Each mask at timestep $t=1,\ldots,T_{max}$ is selected by the left-hand side of the production rule indicated in the 1-hot vector $\+x_t$. 
Given these masks we can compute the decoder's mapping
\begin{align}
p(\+X | \+z) = \prod_{t=1}^{T(\X)} p(\x_t | \+z),
\end{align}
with the individual probabilities at each timestep defined as in Eq.~(\ref{eq:prob-productions}). 
We pad any remaining timesteps after $T(\X)$ up to $T_{max}$ with a dummy rule, a one-hot vector indicating the parse tree is complete and no actions are to be taken.

In all our experiments, $q(\+z | \+X)$ is a Gaussian distribution whose mean and variance parameters are the output of the
encoder network, with an isotropic Gaussian prior $p(\z)\!=\!\mathcal{N}(0,\mathbf{I})$. 
At training time, we sample a value of $\+z$ from $q(\z|\X)$ to compute the ELBO
\begin{align}
\mathcal{L}(\phi, \theta; \X)
&= 
\E_{q(\z|\X)}\left[ \log p_\theta(\X,\z) - \log q_\phi(\z|\X) \right].
\label{eq:elbo-big}
\end{align}
Following \citet{kingma2014auto}, we apply a non-centered parameterization on the encoding Gaussian distribution
and optimize Eq.~\eqref{eq:elbo-big} using gradient descent,
learning encoder and decoder neural network parameters $\phi$ and $\theta$.
Algorithm 2 summarizes the training procedure.

\begin{algorithm}[t]

\label{algo:train}
\begin{algorithmic}[1]

\REQUIRE Dataset $\{\X^{(i)}\}_{i=1}^N$
\ENSURE Trained VAE model $p_{\theta}(\X | \z), q_{\phi}(\z | \X)$
\WHILE{VAE not converged}
	\STATE Select element: $\X \in \{\X^{(i)}\}_{i=1}^N$ (or minibatch)
	\STATE Encode: $\z \sim q_{\phi}(\z | \X)$
	\STATE Decode: given $\z$, compute logits $\mathbf{F} \in \mathbb{R}^{T_{max} \times K}$
	\FOR{$t$ in $[1,\ldots, T_{max}]$}
		\STATE Compute $p_{\theta}(\x_t | \z)$ via Eq.~\eqref{eq:prob-productions}, with mask $\+m_{\x_t}$ and logits $\mathbf{f}_t$ 
	\ENDFOR
	\STATE Update $\theta,\phi$ using estimates $p_{\theta}(\X | \z), q_{\phi}(\z | \X)$, via gradient descent on the ELBO in Eq.~\eqref{eq:elbo-big}
\ENDWHILE
\caption{Training the Grammar VAE}
\end{algorithmic}
\end{algorithm}


\section{Experiments}

We show the usefulness of our proposed grammar variational autoencoder (GVAE)
on two sequence optimization problems: 1) searching for an arithmetic
expression that best fits a dataset and 2) finding new drug molecules. We begin
by showing the latent space of the GVAE and a character variational autoencoder
(CVAE), similar to that of \citet{gomez2016automatic}\footnote{https://github.com/maxhodak/keras-molecules}, on each of the problems.
We demonstrate that the GVAE learns a smooth, meaningful latent space for
arithmetic equations and molecules. Given this we perform optimization in this
latent space using Bayesian optimization, inspired by the technique of
\citet{gomez2016automatic}. We demonstrate that the GVAE improves upon a
previous character variational autoencoder, by selecting
an arithmetic expression that matches the data nearly perfectly, and by finding novel molecules with better drug properties.

\subsection{Problems}
We describe in detail the two sequence optimization problems we seek to solve.
The first consists in optimizing the fit of an arithmetic expression. We are
given a set of 100,000 randomly generated univariate arithmetic expressions
from the following grammar:
\begin{lstlisting}[mathescape,basicstyle=\small,stringstyle=\color{green}]
S $\rightarrow$ S `+' T | S `*' T | S `/' T | T
T $\rightarrow$ `(' S `)' | `sin(' S `)' | `exp(' S `)'
T $\rightarrow$ `x' | `1' | `2' | `3'
\end{lstlisting}
where S and T are non-terminals and the symbol | separates the possible
production rules generated from each non-terminal. By parsing this grammar we
can randomly generate strings of univariate arithmetic equations (functions of $x$) such as the following:
\function{sin(2)}, \function{x/(3+1)}, \function{2+x+sin(1/2)}, and \function{x/2*exp(x)/exp(2*x)}. We limit the
length of every selected string to have at most 15 production rules. Given this
dataset we train both the CVAE and GVAE to learn a latent space of arithmetic
expressions. We propose to perform optimization in this latent space of
expressions to find an expression that best fits a fixed dataset. A common
measure of best fit is the test MSE between the predictions made by a selected
expression and the true data. In the generated expressions, the presence of
exponential functions can result in very large MSE values. For this reason, we
use as target variable $\log(1 + \text{MSE})$ instead of MSE.

For the second optimization problem, we follow \cite{gomez2016automatic} and
optimize the drug properties of molecules. Our goal is to maximize the
water-octanol partition coefficient (logP), an important metric in drug
design that characterizes the drug-likeness of a molecule. As in
\citet{gomez2016automatic} we consider a penalized logP score that takes into
account other molecular properties such as ring size and synthetic
accessibility \cite{ertl2009estimation}. The training data for the CVAE and
GVAE models are 250,000 SMILES strings \cite{weininger1988smiles} extracted at
random from the ZINC database by \citet{gomez2016automatic}. We describe the context-free
grammar for SMILES strings that we use to train our GVAE in the supplementary
material.

\begin{table}[t]
\centering
{\scriptsize
\begin{tabular}{l || l}
\hline
{\normalsize {\bf Character VAE }} & {\small {\bf Grammar VAE}} \\ \hline 
\texttt{\textbf{3*x+exp(3)+exp(1)}}&\texttt{\textbf{3*x+exp(3)+exp(1)}} \\
{\tt\color{darkgray} 2*2+exp(3)+exp(1) } & {\tt\color{darkgray}  3*x+exp(3)+exp(1) } \\
{\tt\color{darkgray} 3*1+exp(3)+exp(2) } & {\tt\color{darkgray}  3*x+exp(x)+exp(1/2) } \\
{\tt\color{red} 2*1+exp3)+exp(2) } & {\tt\color{darkgray}  2*x+exp(x)+exp(1/2) } \\
{\tt\color{darkgray} 2*3+(x)+exp(x*3) } & {\tt\color{darkgray}  2*x+(x)+exp(1*x) } \\
{\tt\color{darkgray} 2*x+(2)+exp(x*3) } & {\tt\color{darkgray}  2*x+(x)+exp(x*x) } \\
\texttt{\textbf{2*x+(1)+exp(x*x)}}&\texttt{\textbf{2*x+(1)+exp(x*x)}} \\ \hline
\texttt{\textbf{3*x+exp(1)+(x+3)}}&\texttt{\textbf{3*x+exp(1)+(x+3)}} \\
{\tt\color{darkgray} 3*x+exp(3)+(x*3) } & {\tt\color{darkgray}  3*x+exp(1)+(x+3) } \\
{\tt\color{darkgray} 3*1+exp(3)+(2*1) } & {\tt\color{darkgray}  2*3+exp(x)+(x) } \\
{\tt\color{darkgray} 3*x+exp(3)+(2*1) } & {\tt\color{darkgray}  2*3+x+(x+3) } \\
{\tt\color{darkgray} 2*1+exp(3)+(x*2) } & {\tt\color{darkgray}  2*3+x+(x/3) } \\
{\tt\color{red} 2*x+exp3)+xx(3) } & {\tt\color{darkgray}  2*2+3+(x*3) } \\
\texttt{\textbf{2*2+3+exp(x*3)}}&\texttt{\textbf{2*2+3+exp(x*3)}} \\ \hline
\texttt{\textbf{x+1+exp(1)+sin(1*2)}}&\texttt{\textbf{x+1+exp(1)+sin(1*2)}} \\
{\tt\color{darkgray} x+1+exp(1)+sin(1*2) } & {\tt\color{darkgray}  x+1+exp(1)+sin(1*2) } \\
{\tt\color{red} 1+3+exp(x)+(i*1) } & {\tt\color{darkgray}  x/1+exp(x)+sin(x*2) } \\
{\tt\color{darkgray} 3+1+exp(2)+(1*1) } & {\tt\color{darkgray}  x/x+sin(x)+exp(x*2) } \\
{\tt\color{darkgray} x+2+exp(x)+(2*3) } & {\tt\color{darkgray}  3*x+sin(x)+(x*3) } \\
{\tt\color{darkgray} x*3+exp(3)+(3*2) } & {\tt\color{darkgray}  3*x+sin(3)+(3*3) } \\
\texttt{\textbf{3*3+sin(3)+(3*3)}}&\texttt{\textbf{3*3+sin(3)+(3*3)}} \\ \hline
\texttt{\textbf{3*x+sin(2)+(x*x)}}&\texttt{\textbf{3*x+sin(2)+(x*x)}} \\
{\tt\color{red} x*1+exp(x)+ex*3) } & {\tt\color{darkgray}  3*x+sin(2)+(x*x) } \\
{\tt\color{red} x*2+exp(x)+ex*x) } & {\tt\color{darkgray}  3*x+sin(2)+(3*x) } \\
{\tt\color{darkgray} x*2+exp(x)+(x*1) } & {\tt\color{darkgray}  3*x+exp(2)+(3*3) } \\
{\tt\color{darkgray} x*3+exp(x)+(x*3) } & {\tt\color{darkgray}  3*x+exp(2)+(3*3) } \\
{\tt\color{darkgray} x*1+exp(x)+(2*2) } & {\tt\color{darkgray}  3*x+exp(2)+(2*2) } \\
\texttt{\textbf{3*x+exp(2)+(2*2)}}&\texttt{\textbf{3*x+exp(2)+(2*2)}} \\ \hline
\end{tabular}}
\caption{Linear interpolation between two equations (in bold, at top and bottom of each cell).
The character VAE often passes through intermediate strings which do not decode to a valid equation (shown in red).
The grammar VAE makes subjectively smaller perturbations at each stage.
}\label{tab:eq-trajectories}
\vspace{-1.5ex}
\end{table}

\begin{figure*}[t]
\centering
\adjincludegraphics[trim={0 0 {0.5\width} 0},clip,width=0.48\textwidth]{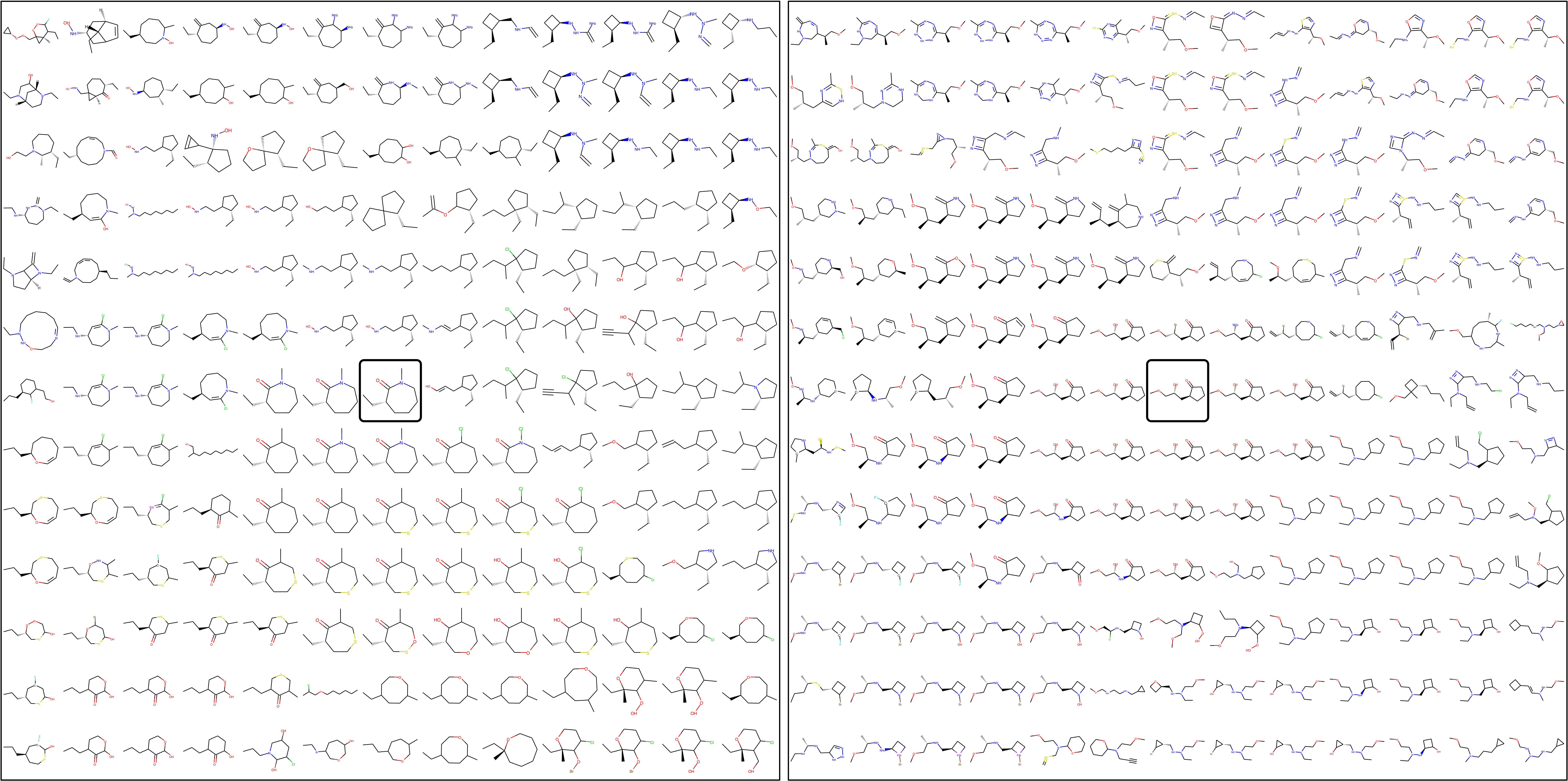}
\hspace{1em}
\adjincludegraphics[trim={{0.5\width} 0 0 0},clip,width=0.48\textwidth]{latent_search_2D.pdf}
\caption{Searching the $56$-dimensional latent space of the GVAE, starting at the molecule in the center.\label{figure.latent_search}}
\end{figure*}

\subsection{Visualizing the latent space}

\paragraph{Arithmetic expressions.}
To qualitatively evaluate the smoothness of the VAE embeddings for arithmetic
expressions, we attempt interpolating between two arithmetic expressions, as in
\citet{bowman2016generating}. This is done by encoding two equations and then
performing linear interpolation in the latent space.  Results comparing the
character and grammar VAEs are shown in Table~\ref{tab:eq-trajectories}.
Although the character VAE smoothly interpolates between the text
representation of equations, it passes through intermediate points which do not
decode to valid equations. In contrast, the grammar VAE also provides smooth
interpolation \emph{and} produces valid equations for any location in the latent
space. A further exploration of a 2-dimensional latent space is shown in the appendix.

\paragraph{Molecules.}
We are interested if the GVAE produces a coherent latent space of molecules. To
assess this we begin by encoding a molecule. We then generate 2 random
orthogonal unit vectors in latent space (scaled down to only search the
neighborhood of the molecules). Moving in combinations of these directions
defines a grid and at each point in the grid we decode the latent vector $1000$
times. We select the molecule that appears most often as the representative
molecule. Figure~\ref{figure.latent_search} shows this latent space search
surrounding two different molecules. Compare this to Figures 13-15 in \citet{gomez2016automatic}. We note that in each plot of the GVAE the latent space
is very smooth, in many cases moving from one grid point to another will only
change a single atom in a molecule. In the CVAE \citep{gomez2016automatic} we do not observe such fine-grained smoothness.


\subsection{Bayesian optimization}
We now perform a series of experiments using the autoencoders to produce novel
sequences with improved properties. For this, we follow the approach proposed
by \citet{gomez2016automatic} and after training the GVAE, we train an
additional model to predict properties of sequences from their latent
representation.  
To propose promising new sequences, we can start from the
latent vector of an encoded sequence and then use the output of this predictor
(including its gradient) to move in the latent space direction most likely to
improve the property. The resulting new latent points can then be decoded into
corresponding sequences. 

In practice, measuring the property of each new sequence could be an expensive
process. For example, the sequence could represent an organic photovoltaic
molecule and the property could be the result of an expensive quantum
mechanical simulation used to estimate the molecule's power-conversion
efficiency \cite{Hachmann_2011}. 
The sequence could also represent a program or expression which may be 
computationally expensive to evaluate.
Therefore, ideally, we would like the optimization process to perform only a
reduced number of property evaluations. For this, we use Bayesian optimization
methods, which choose the next point to evaluate by maximizing an acquisition
function that quantifies the benefit of evaluating the property at a particular
location \cite{shahriari2016taking}.

After training the GVAE, we obtain a latent feature vector for each sequence in
the training data, 
given by the mean of the
variational encoding distributions. We use these vectors and their
corresponding property estimates to train a sparse Gaussian process (SGP) model
with 500 inducing points \cite{snelson2005sparse}, which is used to make
predictions for the properties of new points in latent space. After training
the SGP, we then perform 5 iterations of batch Bayesian optimization using the
expected improvement (EI) heuristic~\cite{jones1998efficient}.  On each
iteration, we select a batch of 50 latent vectors by sequentially maximizing
the EI acquisition function. We use the Kriging Believer Algorithm to account
for pending evaluations in the batch selection process \cite{Cressie_1990}.
That is, after selecting each new data point in the batch, we add that data
point as a new inducing point in the sparse GP model with associated target
variable equal to the mean of the GP predictive distribution at that point.
Once a new batch of 50 latent vectors is selected, each point in the batch is
transformed into its corresponding sequence using the decoder network in the
GVAE.  The properties of the newly generated
sequences are then computed and the resulting data is added to the training set
before retraining the SGP and starting the next BO iteration. Note that some of
the new sequences will be invalid and consequently, it will not be possible to
obtain their corresponding property estimate. In this case we fix the property
to be equal to the worst value observed in the original training data.

\begin{figure}[t]
\centering
\includegraphics[width=0.35\textwidth]{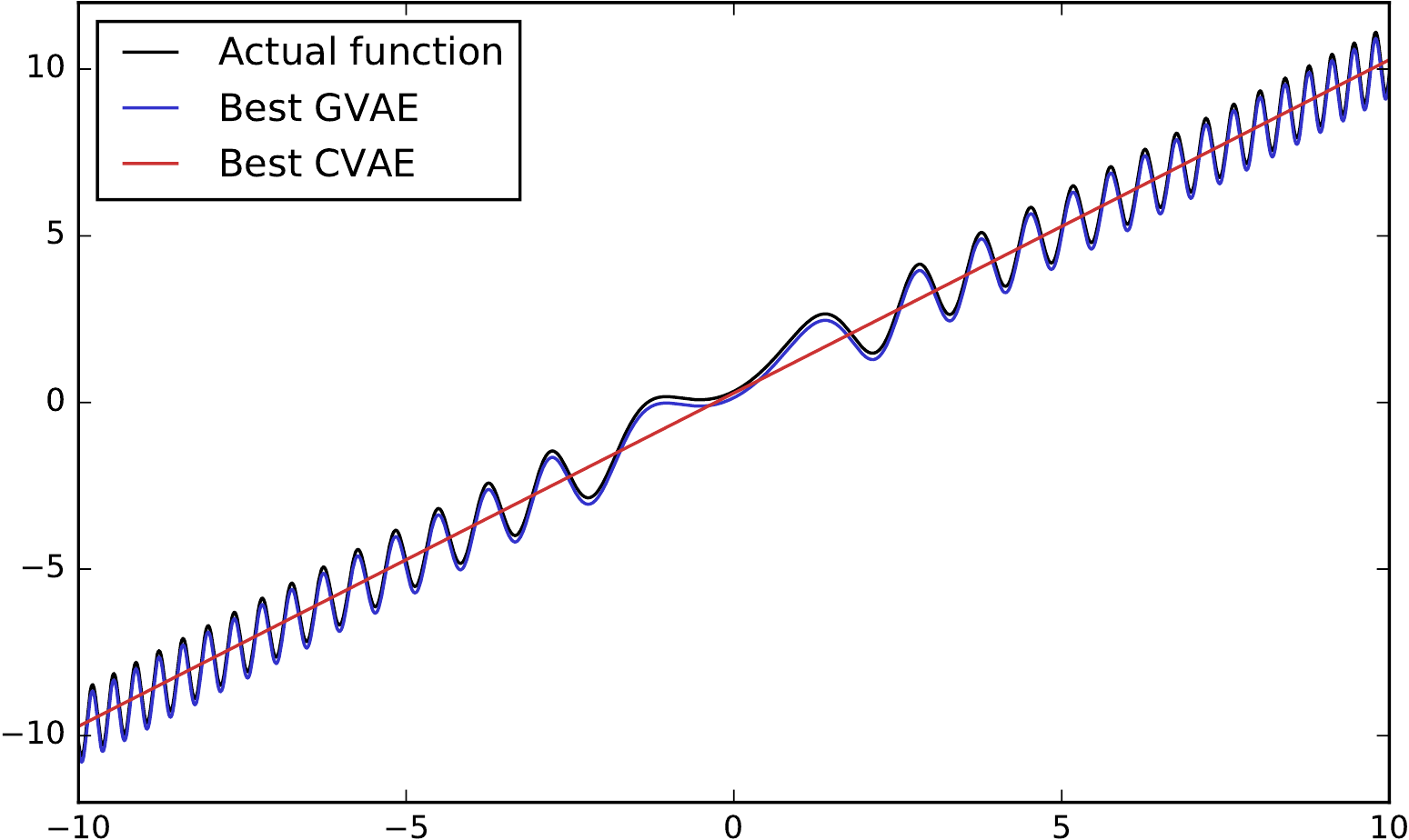}
\caption{Plot of best expressions found by each method}
\label{fig:expressions}
\end{figure}



\begin{table}
\centering
\caption{Results finding best expression and molecule}\label{tab:molecule_statistics}
\begin{tabular}{ccr@{$\pm$}lr@{$\pm$}l}
\hline
\bf{Problem} & \bf{Method}& \multicolumn{2}{c}{\bf Frac. valid} & \multicolumn{2}{c}{\bf Avg. score} \\
\hline
\multirow{2}{*}{Expressions} & GVAE & {\bf 0.99} & {\bf 0.01} & {\bf 3.47 } & {\bf 0.24 } \\
& CVAE & 0.86 & 0.06 & 4.75 & 0.25 \\
\hline
\multirow{2}{*}{Molecules} & GVAE & {\bf 0.31} & {\bf 0.07} & {\bf -9.57 } & {\bf 1.77 } \\
& CVAE & 0.17 & 0.05 & -54.66 & 2.66 \\
\hline
\end{tabular}
\end{table}

\begin{table}
\centering
\caption{Best expressions found by each method}\label{tab:expressions}
\begin{tabular}{llll}
\hline
\bf{Method} & {\bf \#} & {\bf Expression } & {\bf Score}\\
\hline
\multirow{3}{*}{GVAE} & 1 & {\small \function{x/1+sin(3)+sin(x*x)} } & {\bf 0.04 } \\
& 2 & {\small \function{1/2+(x)+sin(x*x)} } & {\bf 0.10} \\
& 3 & {\small \function{x/x+(x)+sin(x*x)} } & {\bf 0.37}  \\
\hline
\multirow{3}{*}{CVAE} & 1 & {\small \function{x*1+sin(3)+sin(3/1)} } & 0.39\\
& 2 & {\small \function{x*1+sin(1)+sin(2*3)} } & 0.40 \\
& 3 & {\small \function{x+1+sin(3)+sin(3+1)} } & 0.40 \\
\hline
\end{tabular}
\end{table}

\paragraph{Arithmetic expressions.}
Our goal is to see if we can find an arithmetic expression that best fits a
fixed dataset. Specifically, we generate this dataset by selecting $1000$ input
values, $x$, that are linearly-spaced between $-10$ and $10$. We then pass
these through our true function \function{1/3+x+sin(x*x)} to generate the true target
observations. We use Bayesian optimization (BO) as described above search for
this equation. We run BO for $5$ iterations and average across 10 repetitions of the process.
Table~\ref{tab:molecule_statistics} (rows 1 \& 2) shows the results obtained. The third column in the table reports
the fraction of arithmetic sequences found by BO that are valid. The GVAE
nearly always finds valid sequences. The only cases in which it does not is
when there are still non-terminals on the stack of the decoder upon reaching
the maximum number of time-steps $T_{max}$, however this is rare. Additionally,
the GVAE finds squences with better scores on average when compared with the CVAE.

Table~\ref{tab:expressions} shows the top 3 expressions found by GVAE and CVAE
during the BO search, together with their associated score values.
Figure~\ref{fig:expressions} shows how the best expression found by GVAE and
CVAE compare to the true function. We note that the CVAE has failed to find the
sinusoidal portion of the true expression, while the difference between the
GVAE expression and the true function is negligible.

\begin{figure}[t]
\begin{center}
\includegraphics[width=0.8\columnwidth]{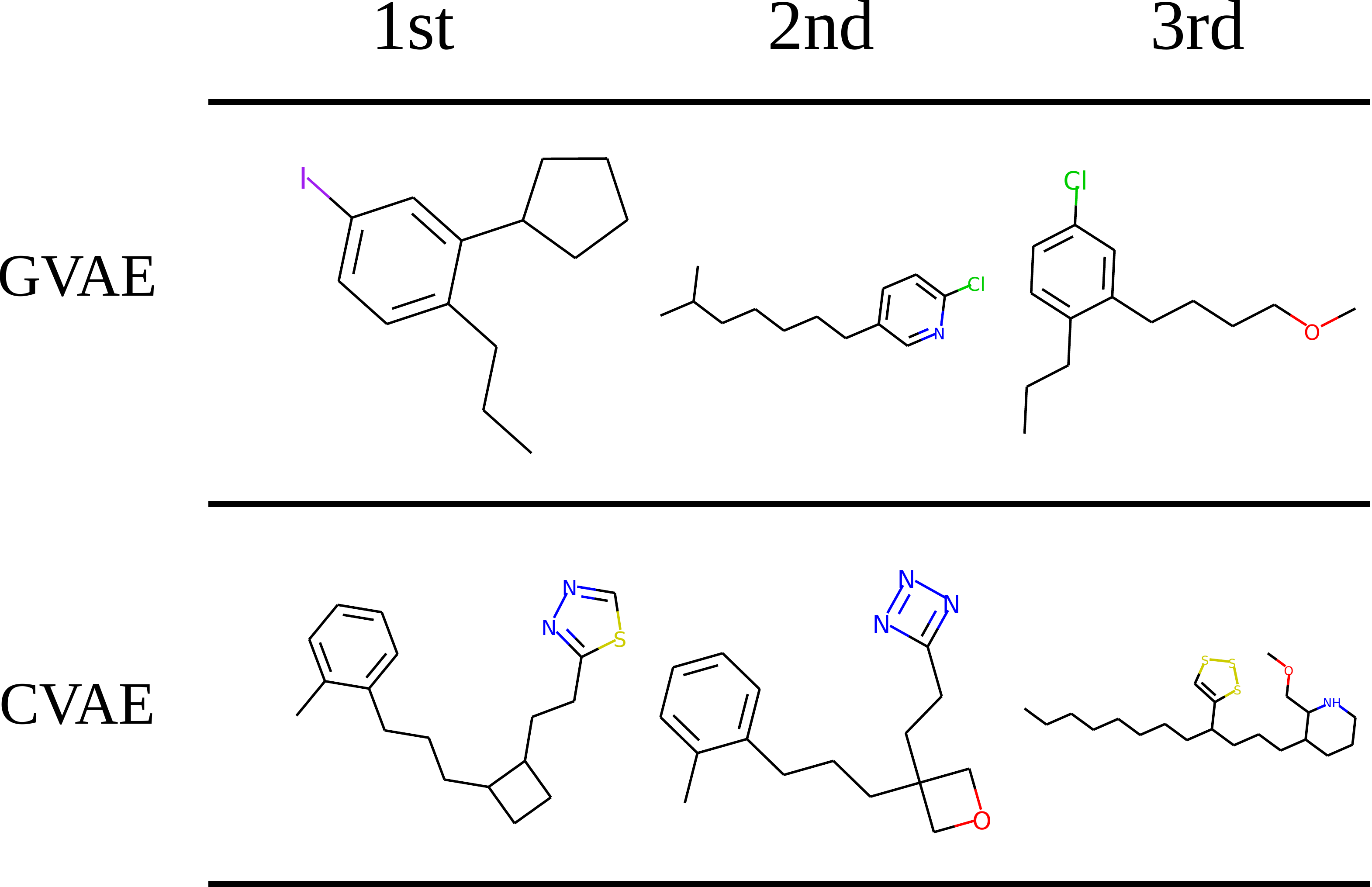}
\caption{Plot of best molecules found by each method.}\label{fig:plot_molecules}
\end{center}
\end{figure}

\begin{table}
\centering
\caption{Best molecules found by each method}\label{tab:molecules}
\resizebox{1.0\columnwidth}{!}{
\begin{tabular}{llll}
\hline
\bf{Method} & {\bf \#} & {\bf SMILE } & {\bf Score}\\
\hline
\multirow{3}{*}{GVAE} & 1 & {\footnotesize CCCc1ccc(I)cc1C1CCC-c1 } & {\bf 2.94 } \\
& 2 & {\footnotesize CC(C)CCCCCc1ccc(Cl)nc1 } & {\bf 2.89} \\
& 3 & {\footnotesize CCCc1ccc(Cl)cc1CCCCOC } & {\bf 2.80}  \\
\hline
\multirow{3}{*}{CVAE} & 1 & {\footnotesize Cc1ccccc1CCCC1CCC1CCc1nncs1} & 1.98 \\
& 2 & {\footnotesize Cc1ccccc1CCCC1(COC1)CCc1nnn1 } & 1.42 \\
& 3 & {\tiny CCCCCCCCC(CCCC212CCCnC1COC)c122csss1 } & 1.19 \\
\hline
\end{tabular}
}
\vspace{-2ex}
\end{table}

\paragraph{Molecules.}
We now consider the problem of finding new drug-like molecules. We
perform $10$ iterations of BO, and average results across $5$ trials.
Table~\ref{tab:molecule_statistics} (rows 3 \& 4) shows the overall BO results. In this problem, the GVAE produces about twice more
valid sequences than the CVAE. The valid sequences produced by the GVAE also
result in higher scores on average.
The best found SMILES strings by each method and their scores are shown in
Table~\ref{tab:molecules}; the molecules themselves are plotted in
Figure~\ref{fig:plot_molecules}.

\subsection{Predictive performance of latent representation}

We now perform a series of experiments to evaluate the predictive performance
of the latent representations found by each autoencoder. For this, we use the
sparse GP model used in the previous Bayesian optimization experiments and look
at its predictive performance on a left-out test set with 10\% of the data,
where the data is formed by the latent representation of the available sequences (these
are the inputs to the sparse GP model) and the associated properties of those
sequences (these are the outputs in the sparse GP model). Table \ref{tab:test_ll} show the average test RMSE and test
log-likelihood for the GVAE and the CVAE across 10 different splits of the data
for the expressions and for the molecules. This table shows that the 
GVAE produces latent features that yield much better predictive performance
than those produced by the CVAE.

\section{Related Work}

Parse trees have been used to learn continuous representations of text in recursive neural network models \citep{socher2013recursive,irsoy2014deep,paulus2014global}. These models learn a vector at every non-terminal in the parse tree by recursively combining the vectors of child nodes. Recursive autoencoders learn these representations by minimizing the reconstruction error between true child vectors and those predicted by the parent \citep{socher2011semi,socher2011dynamic}. Recently, \citet{allamanis2016learning} learn representations for symbolic expressions from their parse trees. Importantly, all of these methods are discriminative and do not learn a generative latent space.

\begin{table}
\centering
\caption{Test Log-likelihood (LL) and RMSE for the sparse GP predictions of penalized LogP score from the latent space}\label{tab:test_ll}
\begin{tabular}{ccr@{$\pm$}lr@{$\pm$}l}
\hline
\bf{Objective} & \bf{Method}& \multicolumn{2}{c}{\bf Expressions} & \multicolumn{2}{c}{\bf Molecules} \\
\hline
\multirow{2}{*}{LL} & GVAE & {\bf -1.320} & {\bf 0.001 } & {\bf -1.739 } & {\bf 0.004} \\
& CVAE & -1.397 & 0.003 & -1.812 & 0.004 \\
\hline
\multirow{2}{*}{RMSE} & GVAE & {\bf 0.884 } & {\bf 0.002 } & {\bf 1.404 } & {\bf 0.006} \\
& CVAE & 0.975 & 0.004 & 1.504 & 0.006 \\
\hline
\end{tabular}
\vspace{-2ex}
\end{table}

Learning arithmetic expressions to fit data, often called \emph{symbolic regression}, are generally based
on genetic programming \cite{willis1997genetic} or other computationally demanding 
evolutionary algorithms to propose candidate expressions \cite{schmidt2009distilling}.
Alternatives include running particle MCMC inference to estimate a Bayesian posterior over parse trees \cite{perov2016automatic}.


In molecular design, searching for new molecules is traditionally done by sifting through large databases of potential molecules and then subjecting them to a virtual screening process \cite{pyzer2015high,gomezOther}. 
These databases are too large to search via exhaustive enumeration, and require novel stochastic
search algorithms tailored to the domain \cite{virshup2013stochastic,rupakheti2015strategy}. \citet{segler2017generating} fit a recurrent neural network to chemicals represented by SMILES
strings, however their goal is more akin to density estimation;
they learn a simulator which can sample proposals for novel molecules, but it is not otherwise used
as part of an optimization or inference process itself.
Our work most closely resembles \citet{gomez2016automatic} for novel molecule synthesis,
in that we also learn a latent variable model which admits a continuous representation
of the domain.
However, both \citet{segler2017generating} and \citet{gomez2016automatic} use character-level models for molecules.




\section{Discussion}
Empirically, it is clear that representing molecules and equations by way of their parse tree
outperforms text-based representations.
We believe this approach will be broadly useful for representation learning, inference, and optimization
in any domain which can be represented as text in a context-free language.

\subsection*{Acknowledgements}

This work was supported by The Alan Turing Institute under the EPSRC grant EP/N510129/1.

\bibliography{grammar}
\bibliographystyle{icml2017}
\vfill
\onecolumn
 \pagebreak
 \appendix
\part*{Appendix}

\section{Grammars for equations and SMILES}

The grammar for the single-variable equations includes 3 binary operators, 2 unary operators, 3 constants, and grouping symbols;
the start symbol is {\tt S}.
Training data for the VAE came by generating 100,000 different equations with parse tree depth less than 7, corresponding to equations which can be produced using up to 15 production rule applications.

\begin{lstlisting}[mathescape,basicstyle=\scriptsize,stringstyle=\color{green}]
S $\rightarrow$ S '+' T | S '*' T | S '/' T | T
T $\rightarrow$ '(' S ')' | 'sin(' S ')' | 'exp(' S ')' | 'x' | '1' | '2' | '3'
\end{lstlisting}

The grammar for SMILES is based on the official \textsc{OpenSMILES} specification \citep{weininger1988smiles},
and starts with {\tt smiles}.

\begin{lstlisting}[mathescape,basicstyle=\scriptsize,stringstyle=\color{green}]
smiles $\rightarrow$ chain
atom $\rightarrow$ bracket_atom | aliphatic_organic | aromatic_organic
aliphatic_organic $\rightarrow$ 'B' | 'C' | 'N' | 'O' | 'S' | 'P' | 'F' | 'I' | 'Cl' | 'Br'
aromatic_organic $\rightarrow$ 'c' | 'n' | 'o' | 's'
bracket_atom $\rightarrow$ '[' BAI ']'
BAI $\rightarrow$ isotope symbol BAC | symbol BAC | isotope symbol | symbol
BAC $\rightarrow$ chiral BAH | BAH | chiral
BAH $\rightarrow$ hcount BACH | BACH | hcount
BACH $\rightarrow$ charge class | charge | class
symbol $\rightarrow$ aliphatic_organic | aromatic_organic
isotope $\rightarrow$ DIGIT | DIGIT DIGIT | DIGIT DIGIT DIGIT
DIGIT $\rightarrow$ '1' | '2' | '3' | '4' | '5' | '6' | '7' | '8'
chiral $\rightarrow$ '@' | '@@'
hcount $\rightarrow$ 'H' | 'H' DIGIT
charge $\rightarrow$ '-' | '-' DIGIT | '-' DIGIT DIGIT | '+' | '+' DIGIT | '+' DIGIT DIGIT
bond $\rightarrow$ '-' | '=' | '#' | '/' | '\'
ringbond $\rightarrow$ DIGIT | bond DIGIT
branched_atom $\rightarrow$ atom | atom RB | atom BB | atom RB BB
RB $\rightarrow$ RB ringbond | ringbond
BB $\rightarrow$ BB branch | branch
branch $\rightarrow$ '(' chain ')' | '(' bond chain ')'
chain $\rightarrow$ branched_atom | chain branched_atom | chain bond branched_atom
\end{lstlisting}

%
%
\section{Network structure}
We briefly overview recent sequence modeling advances which inform our encoder and decoder models. 
An encoder $q_\phi(\z|\+X)$ takes a sequence of $T$ timesteps $\+X = [\x_1, \ldots, \x_T]$ as input and returns a distribution over real-valued vectors $z$,
whereas the decoder $p_\theta(\x|\z)$ takes a real-valued vector $z$ as input and generates
a distribution over sequences themselves $\+X$.
We use the same encoder and decoder as \citet{gomez2016automatic}, inspired by \citet{bowman2016generating}, with a few modifications as described in Section~\ref{sec:method}. In general, we encode the data as sequences of one-hot vectors
and apply a series of one-dimensional convolutions to the sequence data \citep{kalchbrenner2014convolutional}. These are followed by fully-connected layers that predict 
the mean and variance parameters of a Gaussian distribution $q_\phi(\z|\x)$.
To decode, we use recurrent neural network models for sequences \citep{graves2013generating,cho2014learning}, to output discrete probabilities over symbols at each timestep to define $p_\theta(\x|\z)$. For more architecture details see \citet{gomez2016automatic}.


\begin{table}[h!]
\centering
\caption{Reconstruction accuracy and sample validity results.}\label{table.acc}
\begin{tabular}{ccc}
\hline
\bf{Method} & {\bf \% Reconstruct} & {\bf \% Prior Valid } \\
\hline
GVAE & {\bf 53.7} & {\bf 7.2} \\
CVAE  & 44.6 & 0.70 \\
\hline
\end{tabular}
\end{table}

\section{Additional experiments}



%

\paragraph{Molecule reconstruction \& validity.}
We characterize how well the VAE models over molecules are able to reconstruct
input sequences from their corresponding latent representations and to also
decode valid sequences when sampling from the prior in latent space.
Comparisons of full reconstruction accuracy for both the character and grammar
VAEs are shown in
Table~\ref{table.acc}. To compute reconstruction error we start with $5000$ true molecules from a hold-out set. For each molecule we encode it $10$ times, and we decode each encoding $100$ times (as encoding and decoding are stochastic). This results in $1000$ decoded molecules for each of the $5000$ input molecules. We compute the average of these $1000$ decodings that are identical to the input molecule. We then average these averages across all $5000$ inputs to get the percentage of molecules that reconstruct out of the $5,000,000$ attempts. To compute the percentage prior validity we sample $1000$ latent points from the prior distribution $p(\+z) \!=\! \mathcal{N}(0, \mathbf{I})$. We decode each of these points $500$ times and test which of the decoded SMILES strings correspond to valid molecules. We average across all $1000$ points and $500$ trials to yield the percentages in Table~\ref{table.acc}. These results clearly indicate that the proposed GVAE
has higher reconstruction accuracy, and produces a higher proportion of valid sequences when
sampling from the prior.

\begin{figure*}[t]
\begin{center}
\centerline{\includegraphics[width=0.95\textwidth]{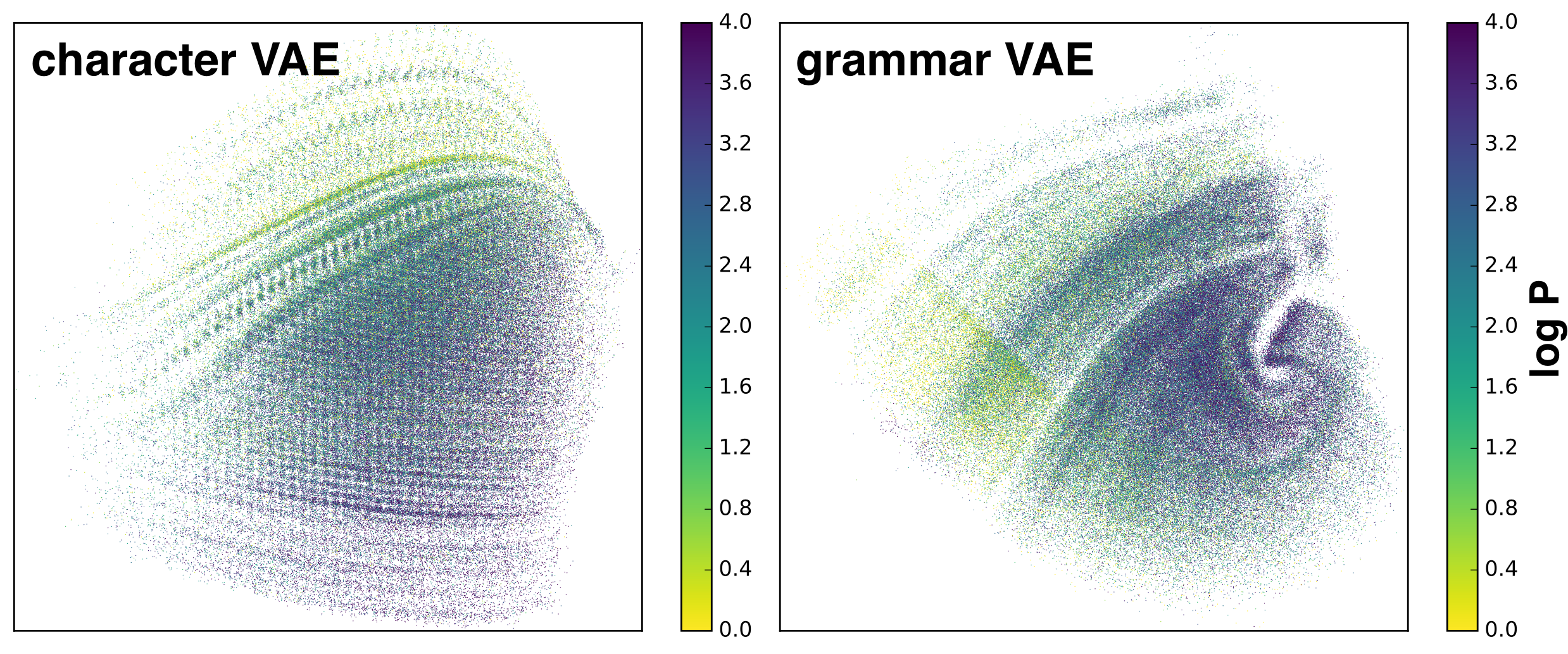}}
\hspace{4em}
\caption{The logP values of a $2$-dimensional character and grammar VAE.
The grammar VAE leads to a low-dimensional latent space which is visually smother with respect to the property of interest. \label{figure.logP}}
\end{center}
\end{figure*}

\paragraph{LogP Visualization.}
To visualize the latent space of the VAEs on molecules with train a CVAE and GVAE on the ZINC dataset \cite{gomez2016automatic} with a $2$-dimensional latent space. We plot the training set colored by the logP values of the molecules in Figure~\ref{figure.logP}. We note that the CVAE seems to have higher logP values (corresponding to molecules with better drug properties) in the lower portion of the latent space. The GVAE on the other hand concentrates molecules with high logP in a small region of latent space. We suspect this makes Bayesian optimization
for molecules with high logP much easier in the GVAE.

\end{document}